%% file: main.tex
\documentclass{article}

\usepackage[preprint]{neurips_2026}
\usepackage{float}  
\AtBeginDocument{\hypersetup{pdftitle={APIFlow-Bench: Measuring Whether Agents Survive Long, Dependent API Workflows}, pdfauthor={Zelin Wan, Arash Nourian, Xiaoxiao Li, Nihar Nandan, Kamalakannan Nandagopal}}}
\makeatletter\renewcommand{\@noticestring}{}\makeatother  

\newcommand{\ANON}{0}
\newcommand{\anonurl}[2]{\ifnum\ANON=1 \texttt{[anonymized for review]}\else\href{#1}{#2}\fi}

\usepackage[utf8]{inputenc}
\usepackage[T1]{fontenc}
\usepackage{hyperref}
\usepackage{url}
\usepackage{booktabs}
\usepackage{amsfonts}
\usepackage{amsmath}
\usepackage{nicefrac}
\usepackage{microtype}
\usepackage{graphicx}
\usepackage{xcolor}
\usepackage{tikz}
\usepackage{pifont}
\usepackage{caption}
\usepackage{subcaption}
\usepackage{multirow}
\usepackage{array}
\usepackage{makecell}
\usetikzlibrary{arrows.meta,positioning,calc,fit,backgrounds,shapes.geometric}

\definecolor{blue1}{HTML}{0072B2}
\definecolor{green1}{HTML}{009E73}
\definecolor{orange1}{HTML}{D55E00}
\definecolor{blkgray}{HTML}{1A1A1A}

\newcommand{\yes}{\textcolor{green1}{\ding{51}}}
\newcommand{\no}{\textcolor{blkgray!40}{\ding{55}}}
\newcommand{\hlf}{\textcolor{orange1}{$\bullet$}}
\newcommand{\bench}{APIFlow-Bench}

\hypersetup{colorlinks=true,linkcolor=blue1,citecolor=blue1,urlcolor=blue1}

\title{\bench: Measuring Whether Agents Survive Long, Dependent API Workflows}

\author{%
  Zelin Wan \quad Arash Nourian  \quad Xiaoxiao Li \quad Nihar Nandan \quad Kamalakannan Nandagopal \\[2pt]
  Postman, Inc. \\
  {\small\texttt{\{zelin.wan, arash.nourian, xiaoxiao.li, nihar.nandan, kamalakannan\}@postman.com}} \\
}

\begin{document}
\maketitle

\begin{abstract}
Tool-using agents are commonly evaluated by a single bit: whether an end-to-end workflow completed. This metric fails to distinguish failures that matter in production, such as expired credentials, malformed payload fields, unverified writes, or correct execution followed by incorrect final delivery. We introduce APIFlow-Bench, a fully auditable benchmark for long-horizon, dependent REST-API workflows that decomposes performance into seven engineering capabilities and requires agents to produce answers supported by the actual call path. In place of mining tasks, we procedurally generate synthetic API worlds forward, subtask by subtask. Each subtask is admitted only when a zero-LLM self-test triad verifies that its grader is load-bearing and an oracle establishes solvability (pass@10 $\geq$ 3). A subsequent adversarial audit identified and fixed six grader exploits. Evaluation is deterministic and provenance-sensitive: a state check traces a mock-minted canary through the API data flow to the response from which the answer is required to originate, while a typed answer card is verified field by field. We release all answer keys and 44,362 unredacted execution transcripts since successful completion depends on evidence available only through the correct call path. Across 19 frontier and open-weight models evaluated under one neutral scaffold, we find the following. (1) Longer dependency chains degrade success, from 93\% on individual subtasks to 74\% on clean 20-subtask chains and 61\% when including the 8\% of chain trials that a model-consensus screen flags as passed by no model. (2) Reliability separates models more than best-case capability does, with best-of-five performance spanning seven points but all-five-of-five reliability spanning 44 points above an approximately seven-point noise floor. (3) The standard independent-error account of compounding failure does not fit the data: observed pass rates on 20-subtask chains are 33 percentage points higher than the product of the corresponding subtask-level pass rates. This gap is not explained by agents simply failing late in execution. On the clean slice, 77\% of failing runs drove the world to the correct final state and failed only at final delivery. 
\end{abstract}

\section{Introduction}

In enterprise systems, consequential work is often performed through API calls embedded in business logic. An agent operating in such environments needs to authenticate, inspect state, interpret failures, repair malformed requests, retry safely, preserve dependencies across calls, and return an output that a downstream system can trust.  

Existing state-based benchmarks, such as $\tau$-bench \citep{taubench2024} and
AppWorld \citep{appworld2024}, appropriately evaluate whether an agent reaches a target world state. However, a single workflow-level pass/fail outcome fails to distinguish an agent that successfully recovers from an authentication error from one that follows an error-free trajectory. In addition, such an outcome conflates an agent that correctly executes the workflow with an agent that reaches the correct backend state but fails to return the required result. 

We introduce APIFlow-Bench, a benchmark for long, dependent REST-API workflows designed to expose this failure surface. APIFlow-Bench evaluates seven capabilities separately: authentication, discovery, schema repair, multistep execution, error recovery, pagination, and statefulness. Each task begins as a realistic service ticket: an initial failing API call, a truthful error signal, and a request to complete a job. Tasks can be composed into cumulative workflows of up to 20 dependent subtasks within one evolving synthetic API world.

Generated tasks provide privacy, scale, controlled variation, and reproducibility, but they introduce a central validity question: are the tasks solvable, correctly graded, and resistant to shortcuts? We therefore treat benchmark validation as a first-class contribution. A generated subtask enters the bank only after passing a zero-LLM self-test triad designed to expose grader failures and an oracle solvability gate evaluated in the same harness used for models. Each assembled chain also passes a two-sided golden replay. Finally, an independent adversarial reviewer attempted to exploit the completed grader. The audit identified six grader exploits, all of which we fixed.

APIFlow-Bench uses deterministic, provenance-gated evaluation. A run has to satisfy two conditions. First, a state check verifies that a mock-minted canary reaches the final workspace through the API data flow required by the task. Second, a typed answer card is checked field by field.

\paragraph{Contributions.}
\begin{enumerate}\setlength{\itemsep}{1pt}
\item \textbf{APIFlow-Bench:} a frozen, fully public benchmark comprising 467 tasks across 13 generated REST API worlds, 241 individual subtasks and 226 cumulative chains of up to 20 subtasks. It evaluates seven capability axes using a seven-tool action space over five typed entity kinds. We release a content-hash-pinned manifest and 44,362 unredacted execution transcripts (\S\ref{sec:bench}).\footnote{Harness and frozen bank: \href{https://github.com/postmanlabs/APIFlow-Bench}{\nolinkurl{github.com/postmanlabs/APIFlow-Bench}}; raw transcripts: \href{https://github.com/postmanlabs/apiflow-bench-transcripts}{\nolinkurl{github.com/postmanlabs/apiflow-bench-transcripts}}; leaderboard: \href{https://www.postman.com/ai/apiflow-leaderboard/overview/}{\nolinkurl{postman.com/ai/apiflow-leaderboard}}.}

\item \textbf{Provenance-gated grading:} a deterministic two-surface evaluator that requires both correct backend state and correct final delivery. The state check traces a mock-minted canary in the final workspace to the API response from which it originates. A typed answer card is then compared field by field (\S\ref{sec:grading}).

\item \textbf{A validation stack for generated benchmarks:} a zero-LLM grader self-test triad, an oracle solvability criterion of pass@10 $\geq$ 3, two-sided golden replay for assembled chains, and an adversarial audit of the completed grader that revealed six exploits missed by automated validation (\S\ref{sec:building}).

\item \textbf{An empirical anatomy of long-horizon API-agent failure:} across 19 models under a common scaffold, we find sharp degradation with chain length, a large reliability gap that best-of-k masks, systematic departures from independent-error predictions, and a predominance of final-delivery failures after otherwise correct state execution (\S\ref{sec:results}).

\item \textbf{A model-consensus screen for cells no model passes:} a practical diagnostic for generated benchmarks based on world$\times$length cells unsolved by a broad model panel. The screen flags 18 of 226 chain cells, or 8.0\% of chain trials. For each flagged cell a reference solution passes the assembled grader (the tasks are solvable), while the cause of the zero pass rate is left open. We report filtered and unfiltered results and document where this heuristic fails (\S\ref{sec:defects}).
\end{enumerate}

\section{Related Work}

\label{sec:related}
\paragraph{Function-call and API-use evaluation:}
Function-calling benchmarks evaluate whether a model selects appropriate tools and constructs valid arguments. API-Bank \citep{apibank2023}, ToolLLM/ToolBench \citep{toolllm2024}, Gorilla \citep{gorilla2023}, and the Berkeley Function Calling Leaderboard (BFCL) \citep{bfcl2024}  have established rigorous settings for function selection, argument construction, and, in some cases, serial, parallel, multi-turn, or multi-step calls.  

APIFlow-Bench addresses a different unit of analysis: the completion of a dependent REST workflow where correct calls may be insufficient.   It measures how call-level competence composes into reliable, stateful workflow execution. 

\paragraph{Stateful interactive environments:} A complementary line of work evaluates agents in environments with persistent state and grades task outcomes instead of matching an action transcript. AppWorld \citep{appworld2024} evaluates interactive coding agents in a controllable world of applications and people, including task success and collateral effects. 
$\tau$-bench \citep{taubench2024}  evaluates customer-service agents that interact with simulated users and domain APIs under policy constraints, comparing final database states to annotated goal states. It also introduced  pass$^k$ for repeated-trial reliability. ToolSandbox \citep{toolsandbox2024} evaluates stateful, conversational tool use, including implicit state dependencies and dynamically assessed intermediate and final milestones. Broader agent benchmarks such as AgentBench \citep{agentbench2024}, WebArena \citep{webarena2024}, OSWorld \citep{osworld2024}, SWE-bench \citep{swebench2024}, and Terminal-Bench \citep{terminalbench2026} evaluate analogous capabilities in web, desktop, software-engineering, and terminal settings.
We share the central premise of this literature: final world state is a necessary component of faithful agent evaluation. Table~\ref{tab:related} (Appendix~\ref{app:related}) tabulates the coverage comparison. APIFlow-Bench differs in granularity and experimental control. It retains state-based evaluation while separately exposing the seven capabilities introduced in Section 1.  

\paragraph{Long-horizon reliability:} Long-horizon tasks reveal failure modes that are often hidden by short, independently sampled instances.  $\tau$-bench's pass$^k$ evaluates the probability that all $k$ independent executions succeed, not whether a single execution does.  APIFlow-Bench adopts repeated-trial evaluation to distinguish reach from reliability: best-of-five performance measures whether a model can succeed, while all-five-of-five measures whether it always succeeds under the same task conditions.

Our controlled chain construction permits a more specific test of horizon effects. We grow dependent subtasks forward within the same synthetic API world, so chain length changes without intentionally changing the tool surface, task distribution, or labeling rule. This design lets us compare observed long-chain success with predictions derived from subtask-level rates.  

\paragraph{Benchmark integrity and generated tasks:} Public benchmarks can become unreliable measurements through training-data contamination, benchmark saturation, underspecified task validity, and evaluators that introduce uncontrolled judgment error.  BIG-Bench \citep{bigbench2023} publishes a canary intended to help model developers exclude benchmark data from training corpora. Deterministic evaluators also avoid known risks of LLM-as-a-judge scoring, including position, verbosity, and self-preference biases documented in recent judge-evaluation work \citep{llmjudge2023}. These concerns motivate our content-hash-pinned manifest, public artifacts, deterministic ranking evaluator, and use of any LLM verification only for non-ranking labels.

Generated interactive benchmarks additionally require evidence that tasks are solvable, graders are sound, and evaluation shortcuts are unavailable. APIFlow-Bench supplies this evidence through the four-safeguard validation stack of \S\ref{sec:building}. We further use model-panel consensus as a diagnostic (i.e., non-conclusive) screen for cells no model passes: a world$\times$length cell unsolved by all models in a broad panel receives manual investigation. This screen identified 18 cells that no model passes in our bank, while our analysis also documents a further such cell that it misses. 

\section{\bench: Interface, Tasks, and Composition}
\label{sec:bench}

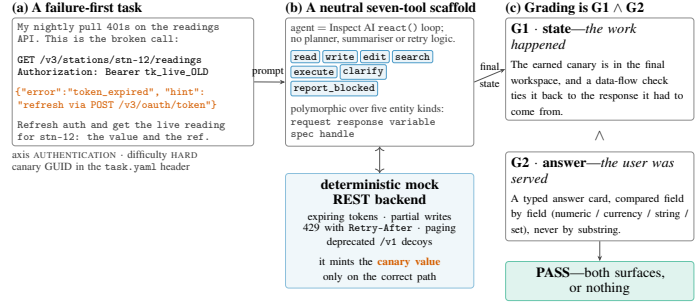
\begin{figure}[t]
\centering
\resizebox{0.65\textwidth}{!}{\input{fig-overview.tex}}
\caption{One \bench{} task, end to end: the \texttt{stn-12} ticket, this paper's running
example. (a)~A failure-first task; (b)~seven polymorphic tools over five typed entity kinds
against a deterministic local mock, under Inspect AI's stock \texttt{react()} loop; (c)~a pass
requires both surfaces: the state check verifies the backend state produced by the calls; the answer card verifies
the answer returned to the user.}
\label{fig:overview}
\end{figure}

\paragraph{The action space is a fixed contract of seven polymorphic tools.}
An agent acts through exactly seven tools (\texttt{read}, \texttt{write}, \texttt{edit},
\texttt{search}, \texttt{execute}, \texttt{clarify}, \texttt{report\_blocked}), each
polymorphic over five typed \texttt{EntityRef} kinds (\texttt{request}, \texttt{response},
\texttt{variable}, \texttt{spec}, \texttt{handle}).  Hence,
protocol differences are parameters, not new tools, and the measured surface stays
identical: the REST bank, GraphQL adapter, and MCP adapter share those seven. \texttt{clarify} and \texttt{report\_blocked} let the agent stop. An abstaining trial forfeits
the ranked metric but is recorded separately, since a benchmark that fails to tell a calibrated
stop from a confabulated answer rewards guessing (\S\ref{sec:discoverable}).

\paragraph{Tasks are failure-first tickets.}
Each task opens in the state developers are in when they ask for help, with something broken,
the offending call inline, and the error carrying a hint. In the \texttt{stn-12} ticket
(Figure~\ref{fig:overview}a), a nightly pull fails with a 401 on an expired token. The hint names the
refresh route, and the task requests the live reading: its value and its reference identifier. Across
the seven axes the same structure repeats.  Table~\ref{tab:axes} (Appendix~\ref{app:axes}) lists the axes. Each
appears at each difficulty level and is scored separately, so a model's profile is a vector.

\paragraph{Worlds, subtasks, solo and chain tasks.} 
The unit of generation is a \emph{world}: one procedurally generated mock-API environment
carrying a multi-segment storyline. Each storyline position is a \emph{segment}; the task
frozen from segment $i$ is its \emph{subtask}, and each subtask also ships as a standalone
\texttt{solo} task. A cumulative \texttt{chain-1to}$k$ task starts from the initial world; its instruction is
one compressed goal centered on segment $k$, and reaching it means re-earning whatever earlier
state the terminal grading gates on; planted identifiers make earlier segments load-bearing
where a coupling exists (coupling is sparse by design; \S\ref{sec:building}). Efficient runs finish
well under one call per nominal segment (median 14 among clean passing chain-20 runs). Grading gates the terminal answer card, the terminal state check, and each
dependency-critical earlier state check (\S\ref{sec:grading}). In the frozen 1.0 bank, 467 tasks
span 13 worlds: 241 solo, 226 chain: eleven full 20-subtask worlds plus two truncated at
lengths 5 and 14 (\S\ref{sec:building}); each of those two also ships one
built-but-unverified solo beyond its truncation ($11{\times}20+5+14=239$ gated segments $+2 =
241$ solos). Solo-versus-chain contrasts
therefore hold the world, tooling, and difficulty labels fixed, while carried state and the
terminal segment's identity vary with $k$. We therefore read pooled trends, not individual lengths (\S\ref{sec:results}).

\section{Building a Verifiable Generated Bank}
\label{sec:building}

Each task is synthetic. No user traffic is mined or replayed. Synthesis adds difficulty
control and a regenerable private split, while a hand-authored bank inherits an author's
judgment that a task is fair and solvable. A generated bank has to demonstrate that its tasks are fair and solvable.

\paragraph{Forward growth through a shared world.}
Initialized once as a state $S_0$, a world is propagated through the chain
(Figure~\ref{fig:pipeline}a, Appendix~\ref{app:pipeline}). Subtask $i$ is proposed against
$S_{i-1}$, the agent workspace plus the mock's accumulated side effects, taken from a run that
\emph{passed}. On acceptance it emits $S_i$. A random seed fixes the axis, difficulty, and
coupling sequences, so that a world regenerates deterministically. A \emph{coupling ledger} forces
cross-subtask dependencies (an early subtask mints a server-generated identifier a later one
can consume), which stops a long chain degenerating into $k$ independent short tasks.

\paragraph{Gate 1: certify the grader is load-bearing prior to trusting it.}
A validator that passes regardless of its input provides no signal. Prior to a subtask reaching any model, a zero-LLM triad
runs against its validator: (1) a \emph{blank exam} (the agent does nothing) needs to
\textsc{fail}; (2) the \emph{answer key} (the reference solution) needs to \textsc{pass}; and
(3) a \emph{sabotaged exam} (each check's evidence, corrupted in place) needs to \textsc{fail}.
The first two conditions are necessary but weak. Many uninformative graders satisfy them. The third is what certifies each assertion load-bearing.

\paragraph{Gate 2: an oracle is required to solve it, in the scoring harness.}
Each subtask is then attempted ten times in parallel by an oracle, in the same harness
that later scores the leaderboard. Since the generator-tier model passed at a rate that left no headroom, we selected the oracle from a lower capability tier. A subtask is admitted only
at $\text{pass@}10 \geq 3$. Otherwise,
a reviser model that shares the generator's family (a fact \S\ref{sec:limits} treats as part
of the entanglement) repairs it in place using the failed oracle transcripts as evidence, then the subtask is torn
down and re-proposed. Shipped defaults allow three repair
rounds and a single re-proposal.  If it
fails to clear, the chain is \emph{truncated at the last solid subtask} instead of being padded
with an uncertified subtask or made easier. Two of thirteen worlds terminate this way, at five
and fourteen subtasks. We ship them short, and hence the \emph{headline slice} of
\S\ref{sec:results} is the eleven full-length worlds. Scope notes: the gate is applied to each subtask (assembled \texttt{chain-1to}$k$ tasks are
\emph{not} oracle-solved end to end, a gap \S\ref{sec:defects} shows to matter). The
oracle's generous message budget means it certifies solvability without ensuring difficulty parity.

\paragraph{Gate 3: two-sided golden replay.}
Once assembled, a chain is replayed deterministically. The forward replay of its reference
solution is required to \textsc{pass}. A negative replay, applied wherever a later subtask depends on an
earlier one's result, is required to \textsc{fail}: the terminal segment's reference alone, run from the initial world state $S_0$ with segments $1, \ldots, k{-}1$ skipped. Static checks run alongside. A seed-workspace value equal to an answer-card value (the
\emph{leak predicate}) blocks publication outright. Of the 467 shipped tasks, 465 are
replay-verified. The two exceptions are documented in the release manifest.

\paragraph{Gates need to be discoverable.}
\label{sec:discoverable}
The cheapest way to make a generated task look hard is to hide an undocumented requirement, a
mandatory header no error message ever reveals. We forbid it on incentive grounds. Against a
hidden trap, a reckless model guesses and sometimes gets lucky, while a calibrated model correctly
concludes it is blocked, calls \texttt{report\_blocked}, and is marked wrong. The trap inverts
the thing being measured. Each gate on the success path needs to be learnable from the served spec or
a truthful error hint. Harder tasks demand more inference, not more guesswork. Under this rule and Gate~2, stacking obstacles failed to manufacture difficulty (discoverable ones are satisfied by inference. Undiscoverable ones drove the oracle pass rate to zero). This implies that a low pass rate alone fails to distinguish ``hard'' from ``flagged''
(\S\ref{sec:defects}). What restores headroom is length (\S\ref{sec:results}).

\section{Grading: Certify the Labor, Then the Delivery}
\label{sec:grading}

A substring check fails in both directions, rejecting \texttt{48,210} for \texttt{48210} and
accepting a lucky coincidence, so \bench{} does not grade text. A run passes only if two
independent surfaces hold.

\textbf{G1 (state): the required calls are executed.} Nearly all tasks have a \emph{canary value} that is
distinctive, hard-coded in the mock backend, reachable only along the correct path, and
different on each decoy: the deprecated \texttt{/v1} route serves other numbers, the summary
endpoint a rounded one. On the \texttt{stn-12} ticket the canary is the live reading, served
only on the \texttt{/v3} route under the refreshed token. G1 requires that this value appear in
the agent's final workspace \emph{and} that a data-flow check tie it to the response it needed to
come from. Possessing it is evidence of having made the real call. (28 of the 241 solo tasks ship with an empty \emph{state signature} (the task-specific list of state requirements the state check verifies) and are gated by the
answer card and structural checks alone. Chains inherit their segments' checks.)

\textbf{G2 (answer): the final answer matches the reference.} The agent declares its final answer as an
\emph{answer card}, compared field by field with a typed comparator (numeric, currency, string,
set). Format is normalized, leaving a correct answer intact under punctuation differences. Structure remains
strict, failing a run that buries the number in prose or drops the reference identifier, even when the
state check passes.
 Checkpoints (such as whether the agent applied backoff after an HTTP 429 response) are diagnostics that do not determine a pass. A parallel LLM verifier reviews units, decimals, and provenance. The verifier can only overturn a pass to a fail and is advisory in this release: the published board is
the deterministic grader's alone.

\paragraph{Provenance gating enables release of answer keys and transcripts.}
Since a pass is gated on evidence an agent can only earn through the real call path, a leaked
expected answer alone fails to let a model pass.  Expected
answers, segment-level reference solutions (239 of 241 solo tasks, chains compose their segments'
references), and all 44{,}362 transcripts ship unredacted. Gating removes the guessing exploit
\emph{within} a run. It fails to prevent training on released solutions, so the release
shares one contamination horizon (\S\ref{sec:limits}). Provenance gating delivers
results and grading that third parties can check (the generation gates' internal logs remain unreleased).

\paragraph{Adversarial audit of the grader.}
 An independent adversarial reviewer attempted to exploit the completed grader by constructing a correct solution it wrongly rejects and a wrong answer it wrongly accepts, each confirmed against the live mock. Six attacks succeeded (Table~\ref{tab:audit},
Appendix~\ref{app:audit}), each involving the grader scoring something other than the work the model
earned. Each fix is re-verified by running the exploit (now required to fail) and the correct solution (required to pass). The audit was one reviewer's single pass, and its attempt coverage is
uninstrumented. The six exploits touch four of the seven axes. No exploit was found on the remaining three axes. This does not establish that they are exploit-free. All six exploits were implementation errors instead of flaws in the grading design, indicating that design-level gates alone fail to validate an implementation. 

\section{Experimental Setup}
\label{sec:protocol}

We evaluate 19 frontier and open-weight models on the frozen 1.0 bank: 467 tasks $\times$ 5
epochs for each model, temperature 1.0, under Inspect AI's stock \texttt{react()}
loop \citep{inspect2024} with no custom planner, summarizer, or retry logic. The panel spans providers, price tiers, and
open- versus closed-weight models. The panel is a slate, not a census: API
access and serving stability are the binding constraints on membership, and no model is dropped
for its score. (One further
frontier model is run and left unranked, because an API-layer safety filter declined
87.8\% of its runs. The model passed 90.9\% where allowed.) Of the 44{,}365-cell grid, 44{,}362 trials ran and 44{,}343 completed without
harness error. We report results based on the completed trials.

Since five epochs of a given chain are five draws from the same world, all confidence intervals are
90\% cluster bootstraps over worlds (2{,}000 resamples), keeping each world's epochs together.
Treating the headline slice's 55 runs of each model as
independent would understate the interval by roughly half. At a true $n$ of eleven worlds the
intervals are wide and all overlap (Figure~\ref{fig:ci}). This implies that the defensible statements are about
pooled levels and the well-powered full-bank statistics of \S\ref{sec:reliability}.

\section{Screening the Bank for Cells No Model Passes}
\label{sec:defects}

We screen the bank with the panel of \S\ref{sec:protocol} prior to reading model
results. A low pass rate alone fails to separate hard from flagged
(\S\ref{sec:discoverable}). On a
wide panel, that has a constructive converse: if each model fails a
world$\times$length cell across all attempts, the cell is consistent both with genuine difficulty and with an instruction that omits a requirement the grader checks. The 19 models span roughly seven independent
training pipelines, and a cell's 95 trials are 19 clustered draws (\S\ref{sec:protocol}). With models as
units, zero passes is consistent with a model-level probability of passing the cell at least once
in five epochs of up to 14.6\% (one-sided 95\% exact binomial, 11.4\% at the paper's usual
90\%). Treating the roughly seven independent pipelines as the unit widens that bound to 34.8\% at
95\%. Hence, the screen nominates and does not prove. Positives require root-cause analysis.

Across the 226 chain cells, 18 have zero passes across all 19 models (1{,}710 of
21{,}455 chain trials, 8.0\%), falling in two families (\texttt{v56-w08} at lengths 13--18 and 20,
\texttt{v56-w10} at 10 and 12--20) plus a single cell of a third (\texttt{v56-w04} at 11).
For each of these cells the assembled grader checks mid-chain steps that the compressed one-goal instruction does not state, yet the assembled chain's forward golden replay passes: a reference solution exists and clears the grader, so the tasks are solvable. Whether the zero pass rate reflects this gap between the compressed instruction and the assembled grader or the capability ceiling of the current models, we do not adjudicate here (see \S\ref{sec:limits}).
No gate could catch it, since Gate~2 runs at the subtask level, prior to assembly, and golden replay
composes segment-level references, each of which passes. The screen, running after assembly,
catches the all-zero cells the unit-level pipeline structurally misses.

Three caveats bound the method. First, 17 of the 18 flagged cells sit in the two
families where signatures gate mid-chain segments. The screen fired where enforcement was
strictest, implying that a correctly built mid-chain-gating chain may also score low. Only
\texttt{w04}@11 is flagged in isolation. Second, the threshold has known escapes:
\texttt{w16}@19 (documented same pattern) and \texttt{w08}@19 (inside that family's length
span but unflagged) each pass 1 of 95. Hence, we read those cells as \emph{near-}zero-pass.
\texttt{w08}@19 and nine further 1--3-pass cells remain unadjudicated. Several show the
dropped-requirement signature at solo-implied rates far above their chain rate, and one solo task
is a zero-pass cell, which makes the solo-clean assumption approximate. Third,
since we performed root-cause analysis only on the screen's positives, we make no specificity claim. A controlled-injection
study measuring precision and recall is the natural next step.

We leave both consequences unrepaired, since the frozen bank needs to stay
reproducible.
\paragraph{Cells no model passes compress aggregate pass rates but do not reorder models.} All models fail the flagged cells. Pooled
chain pass reads 77.0\% with the flagged cells, 83.7\% without. The headline slice moves from
43.6--72.7\% to 53.3--88.9\% (nine worlds), gpt-5.5 leading, deepseek-v4-flash tied
second.

\paragraph{Axis-level slices should not be read raw.} Two of the three \textsc{authentication} worlds
at length 20 are flagged (0.0\%). The survivor passes 17.9\%. Read raw, the slice suggests a 6\% pass rate for authentication at length 20, although most
of that figure reflects the flagged cells instead of task difficulty. 

\section{Results}
\label{sec:results}

\begin{figure}[tb]
\centering
\begin{subfigure}[b]{0.40\textwidth}
  \centering\includegraphics[width=\textwidth]{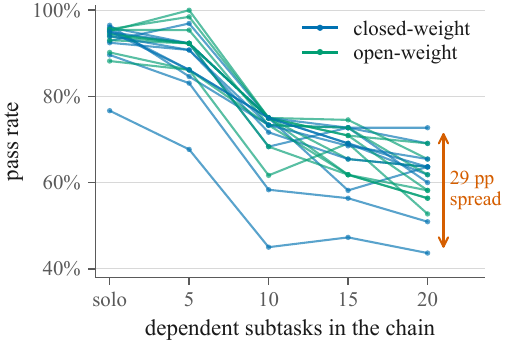}
  \caption{One line per model, five task sets.}
  \label{fig:spread}
\end{subfigure}\hfill
\begin{subfigure}[b]{0.485\textwidth}
  \centering\raisebox{2mm}{\includegraphics[width=\textwidth]{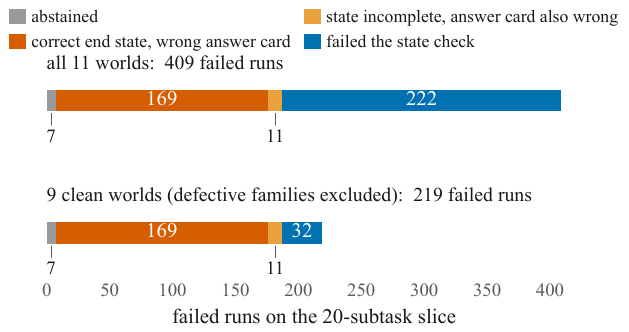}}
  \caption{Failures on the 20-subtask slice.}
  \label{fig:failure}
\end{subfigure}
\caption{(a)~Chain length collapses the level: 93\% pooled on solo, 74\% on clean
20-subtask chains (61\% with the flagged cells). Read the
right-edge fan-out with care: chain-20 rests on eleven worlds and every interval overlaps
every other (Figure~\ref{fig:ci}); Appendix~\ref{app:full} tabulates every point. (b)~Failures
with the screened families of \S\ref{sec:defects} included (top) and excluded (bottom): on
clean worlds the dominant category is final-delivery failure after otherwise correct state execution.}
\end{figure}

\subsection{Chain Length Collapses the Level and Takes the Bank off the Ceiling}
\label{sec:spreadsec}

Chain length is the only parameter that increases difficulty without reducing the
discoverability of task information (\S\ref{sec:discoverable}). Pooled over the panel, pass rates fall from 92.9\% on solo tasks
to 74.4\% on the nine clean 20-subtask chains, 60.9\% with the screened cells
included, at a fixed world set, tool surface, and difficulty labeling
(Figure~\ref{fig:spread}). On solo tasks the bank is at the
ceiling: 18 of 19 models land between 88\% and 97\%, only gpt-5.4-mini below at
77\%. 

 The spread of point
estimates grows from 19.8 points (solo) to 29.1 (chain-20, 8.3 to 21.8 excluding the one low
outlier). However, chain-20 rests on eleven worlds against solo's 241 tasks, and on a log-odds scale,
which undoes the ceiling's compression, the spread does not widen at all (2.1 solo, 1.2
chain-20). At this bank size, model separation is unresolvable here: all
cluster-bootstrap intervals overlap each other (\S\ref{sec:protocol}). The \emph{level
collapse} is the robust chain-length effect. Models separate with headroom in reliability
(\S\ref{sec:reliability}) and failure anatomy (\S\ref{sec:failures}, \S\ref{sec:composition}).

Each accounting leaves closed- and open-weight models interleaved at all lengths and the descent non-monotone in $k$
(Figure~\ref{fig:chainlen}, Appendix~\ref{app:full}). 

\subsection{What the Headline Slice Supports}

Figure~\ref{fig:ci} gives the headline slice, scoped to the 19-model launch panel of the
July 22, 2026 release (the public board has since grown). In this slice, we observed: (1) the point estimates order gpt-5.5 first at 72.7\%, deepseek-v4-flash (a cheap open-weight model)
tied with claude-opus-4-8 for second, and glm-5p2 fourth with qwen3p7-plus; and (2) all intervals
overlap.  The launch blog mislabels qwen3p7-plus as
open-weight. Qwen's ``-plus'' tier is API-only, and both qwen models are classified closed here.
Price sharpens the interleaving (Figure~\ref{fig:pareto}): below \$0.01 a trial the frontier
is entirely open-weight, and gpt-5.5's $+3.6$-point edge costs $\sim$$22\times$ more
(pinned list prices).

\subsection{Long-Horizon Failures Concentrate at Final Delivery After Correct State Execution}
\label{sec:failures}

Of the 20-subtask slice's 409 failing runs, 190 sit on the two screened families.
Roughly half of a full-slice failure decomposition would therefore reflect these screened cells, not model behavior.
Figure~\ref{fig:failure} decomposes both. The analysis below uses the clean subset ($n{=}219$ of 855).

Abstention is nearly absent. Only seven of 219 failures (3.2\%) invoked \texttt{clarify} or
\texttt{report\_blocked}, defensible on clean tasks, where each gate is discoverable by
construction (\S\ref{sec:discoverable}). The other 212
submitted an answer and failed at least one check: 180 failed the typed answer card, and 169 of those
(77\% of all clean-slice failures) failed \emph{only} the answer card, with each
state check green. In these runs the agent drove the world to the correct final state, then
delivered something the typed comparator rejected: a wrong value, missing field, or undeclared
card (the released transcripts support the finer split). The category is world-concentrated (Figure~\ref{fig:heatmap}). Two worlds hold 167 of the 169:
90 in \texttt{w05}, where the cell passes 4.2\%, and 77 in \texttt{w01}, 17.9\%. Hence, the lost
delivery is a property of specific world--task shapes, not a uniform tax. These runs also persist
to the end. The median failing run on this slice makes 15 tool calls (16 among the
answer-card-only failures, maximum 168): they perform the work and lose the thread at
the delivery.

For calibration, the flagged cells add one further observation. Across this slice's 190
failing runs on the flagged tasks, none abstained. Eight of all 1{,}710
flagged-cell trials bank-wide abstained, six of them from a single small model. Facing a flagged
task, models nearly always submit a wrong answer and do not abstain. 

\subsection{The Leaderboard Ranking Primarily Reflects Consistency}
\label{sec:reliability}

\begin{figure}[tb]
\centering
\begin{subfigure}[b]{0.445\textwidth}
  \centering\includegraphics[width=\textwidth]{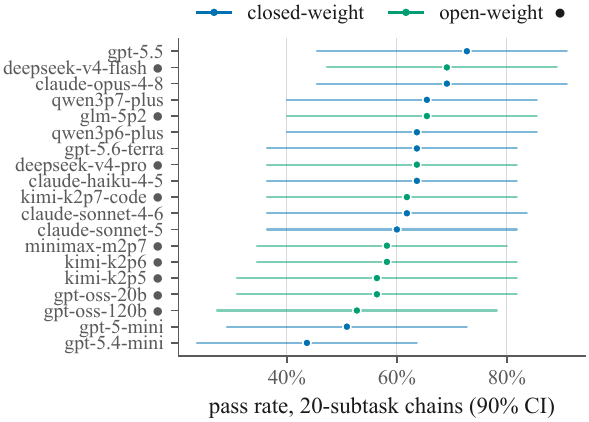}
  \caption{20-subtask slice, 90\% cluster-bootstrap CIs.}
  \label{fig:ci}
\end{subfigure}\hfill
\begin{subfigure}[b]{0.445\textwidth}
  \centering\includegraphics[width=\textwidth]{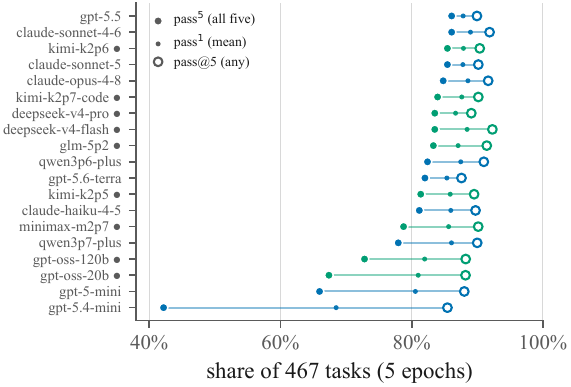}
  \caption{Reliability over five epochs, all 467 tasks.}
  \label{fig:reliability}
\end{subfigure}
\caption{(a)~The headline slice: 1{,}045 ranked runs; every interval overlaps every other, so
the figure supports an ordering of point estimates but does not establish separation between models. (b)~Hollow dot pass@5, small dot pass$^1$, filled dot
pass$^5$ (cells with fewer than five completed epochs excluded): best-of-five compresses the
panel into seven points; all-five-of-five spreads it across 44, a consistency ranking.}
\end{figure}

Five epochs for each task separate two things a single pass rate conflates: whether a model
\emph{can} do a task, and whether it does it \emph{each time}. Figure~\ref{fig:reliability}
reports, for each model over all 467 tasks, pass@5 (any epoch passed), pass$^1$ (the mean), and
pass$^5$ (all five passed), the reliability statistic of \citet{taubench2024}. We exclude tasks
with fewer than five completed epochs (466--467 for each model, 449 for qwen3p7-plus).

In Figure~\ref{fig:reliability}, we observed: (1) best-of-five compresses the whole panel into seven
points (85.4--92.3\%), i.e., after five attempts nearly all tasks are within reach of even the
weakest model; and (2) all-five-of-five spreads the same panel across 44
points (42.2--86.1\%), double the 20.5-point pass$^1$ spread. A binomial null (19 models at the panel-mean pass$^5$ over 467 tasks, max-minus-min across
2{,}000 draws) puts the noise floor near seven points. World-level clustering raises it, and the
observed pass$^5$ spread dwarfs either. The leaderboard is therefore mostly a
\emph{consistency} ranking. Consistency is where small models pay: gpt-5.4-mini drops 26.3
points from pass$^1$ to pass$^5$ and gpt-5-mini 14.6, against 1.8--3.8 for the frontier tier.
The instructive middle case is deepseek-v4-flash: the highest pass@5 on the panel (92.3\%) with a
mid-pack consistency drop, a frontier-class ceiling where the remaining gap is reliability.  A best-of-$n$ wrapper masks nearly the whole capability gap at
$n{=}5$, while an agent wired to side-effecting APIs does not get free retries. Hence, pass$^5$ is the
operationally relevant statistic, and a single-epoch benchmark misses it.

\subsection{Long-Horizon Failure Is Not the Product of Independent Steps}
\label{sec:composition}

Since solo and chain tasks freeze from the same accepted segments, each chain has a natural
null model. If segment $i$ failed independently at its solo pass rate $p_i$ (estimated for each
model and world, so between-model and between-world heterogeneity does not manufacture the
gap), chain-$k$ pass would be $\prod_{i \le k}\,p_i$. For gpt-5.5, a 94.9\% mean solo rate
predicts $0.949^{20}{\approx}35\%$. We observe 72.7\%. Figure~\ref{fig:composition} (Appendix~\ref{app:compfig}) plots that
null against observed chain pass rates on the nine full-length worlds outside the two screened
families (flagged cells w04@11 and w16@19 remain as dips). At $k{=}2$ the pooled curves agree
(84.3\% vs.\ 84.4\%, though large world-level residuals cancel underneath). By $k{=}20$ they
have diverged by 33 percentage points (74.4\% observed vs.\ 41.4\% predicted).

Two mechanisms plausibly produce the gap: the first is a grading artifact, the second genuine competence. First, chain grading is
selective: a \texttt{chain-1to}$k$ task is graded on the final subtask's answer and state plus
dependency-critical earlier checks (\S\ref{sec:bench}). A chain run need not re-earn each
intermediate diagnostic. Second, within-session competence is strongly correlated: a model that handles a world's early
segments tends to handle its later ones, so that segment-level failures rarely multiply. They concentrate
on segments that interact badly with accumulated state, and on the final delivery
(\S\ref{sec:failures}).

Hence, the independent-step account of ``compounding error'' is inconsistent with the data for these agents under chain grading. Length does not tax a chain multiplicatively. Length
samples more of the failure surface in a given task, raising the chance of meeting the segment or
delivery that breaks the run. The difficulty comes from \emph{exposure} instead of hidden
information (\S\ref{sec:spreadsec}). Each gate stays discoverable.

\section{Conclusion, Limitations, \& Future Work}
\label{sec:limits}

\paragraph{Key findings.}
 Three key findings: the level collapse (93\% to 74\% clean); reliability separation (7 vs.\ 44 points); long-horizon failure that is correlated rather than per-step (partly a selective-grading effect). A model-consensus screen flags the cells no model passes, and every aggregate is reported with and without them. Artifacts are public (Apache-2.0, Appendix~\ref{app:datasheet}).

\paragraph{Small $n$ and entanglement.} The headline slice is 1{,}045 runs but eleven sampling units, so Figure~\ref{fig:ci}'s intervals are wide, all overlap, and the top reads as a tie. The second is entanglement. One frontier
family generated all worlds, a sibling served as solvability oracle, a same-family reviser
repaired gate failures (\S\ref{sec:building}), and that family's models are ranked. Mitigations: deterministic validators and golden replay decide pass/fail; the LLM verifier is non-ranking; the gate filters for \emph{solvable}, not \emph{oracle-easy}. A generator-family advantage remains possible: we ran no cross-family re-gating control and retained no gate yield statistics. The headline leader comes from another family.

\paragraph{Scope.} Deterministic mocks trade realism for replayability and difficulty control. A high score means
surviving modeled failure modes under the seven-tool contract, not running arbitrary production systems. The 1.0 bank is REST only (GraphQL and MCP adapters ship without task banks); the panel is the 19-model slate of \S\ref{sec:protocol}. 

\paragraph{Contamination.} All scores predate the public release, so are uncontaminated by it. Later runs can be contaminated (each task carries a GUID
canary \citep{bigbench2023} and a private held-out split is the intended anchor). Future work: more worlds and longer chains, a cross-family re-gating control, and, for the cells no model passes (\S\ref{sec:defects}), methods that preserve requirement discoverability under compression or grading that gates only what the served instruction states.

\clearpage
\bibliographystyle{plainnat}
\bibliography{refs}

\appendix

\section{The Independent-Step Null Model}
\label{app:compfig}
\begin{figure}[H]
\centering
\begin{minipage}[c]{0.45\textwidth}
\includegraphics[width=\textwidth]{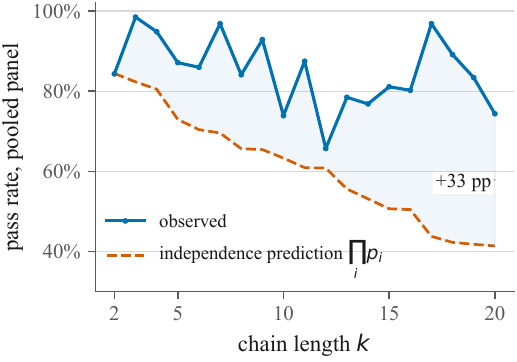}
\end{minipage}\hfill
\begin{minipage}[c]{0.52\textwidth}
\caption{If subtasks failed independently at their solo rates, chain-$k$ pass would be
$\prod_i\,p_i$ (dashed); the observed rate (solid; nine full-length worlds outside the
screened families) sits far above it, and the gap grows with $k$.}
\label{fig:composition}
\end{minipage}
\end{figure}

\section{Overview of the Generation Pipeline}
\label{app:pipeline}

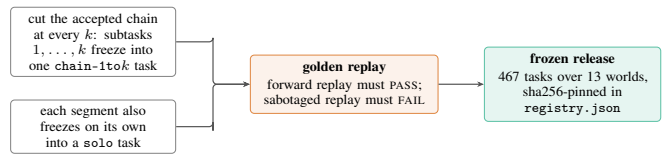
\begin{figure}[H]
\centering
\resizebox{0.78\textwidth}{!}{\input{fig-pipeline.tex}}
\caption{The generation pipeline. (a)~A world is grown forward: each subtask is seeded from the
verified end state of the last, and a coupling ledger forces cross-subtask dependencies that block a
prefix-only shortcut. (b)~Two gates are cleared before models are scored (i.e., a zero-LLM
self-test triad that certifies the grader is load-bearing, and an oracle solvability gate run in the
scoring harness itself). (c)~The accepted chain is cut at every $k$, replayed forward and
sabotaged, and pinned by content hash.}
\label{fig:pipeline}
\end{figure}

\section{Prior-Benchmark Coverage Matrix}
\label{app:related}

\begin{table}[H]
\centering\small
\caption{Where prior benchmarks land on the API-engineering taxonomy. \yes{}~scored as a
first-class axis, \hlf{}~exercised but not separately graded, \no{}~not covered. Marks are our
reading of each benchmark's published task taxonomy, not a quality judgment. Each of these
benchmarks covers part of the picture well, and AppWorld in particular grades pagination and end state.
However, the decomposition itself is rare, and two of the failure modes practitioners encounter most often (i.e., auth refresh and schema repair) are the least targeted.}
\label{tab:related}
\setlength{\tabcolsep}{4.5pt}
\begin{tabular}{@{}lccccc@{}}
\toprule
Benchmark & \makecell{stateful\\chain} & \makecell{end state /\\side effect} & \makecell{auth\\refresh} &
\makecell{pagination\\to end} & \makecell{schema\\repair} \\
\midrule
AppWorld \citep{appworld2024}       & \yes & \yes  & \no  & \yes & \hlf \\
$\tau$-bench \citep{taubench2024}   & \yes & \yes  & \no  & \no  & \no   \\
BFCL v3 \citep{bfcl2024}            & \yes & \hlf & \no  & \no  & \no   \\
SWE-bench \citep{swebench2024}      & \yes & \yes  & \no  & \no  & \no   \\
\midrule
\bench{} (this work)                & \yes & \yes  & \yes & \yes & \yes  \\
\bottomrule
\end{tabular}
\end{table}

\section{The Seven Capability Axes}
\label{app:axes}

\begin{table}[H]
\centering\small
\caption{The seven capability axes, each generated at three creation-time difficulty levels and
composed forward into chains. Pass rates pool the 19 models and all task lengths on the frozen
1.0 bank, where $n$ is trials. The third column excludes the 18 flagged cells of
\S\ref{sec:defects}. Chain tasks carry a single generation-time axis annotation (the chain's
dominant capability, labeled at assembly). Hence, each trial contributes to exactly one row.}
\label{tab:axes}
\setlength{\tabcolsep}{4pt}
\begin{tabular}{@{}llrrr@{}}
\toprule
Axis & What it tests & Pass & Pass (excl.\ flagged) & $n$ \\
\midrule
\textsc{multistep} & carry state across calls & 91.2 & 92.6 & 6{,}457 \\
\textsc{schema} & repair a rejected body & 88.9 & 92.0 & 5{,}604 \\
\textsc{statefulness} & confirm the side effect & 88.4 & 90.8 & 7{,}215 \\
\textsc{pagination} & walk to the last page & 87.8 & 89.6 & 4{,}840 \\
\textsc{error recovery} & back off and retry & 84.6 & 89.3 & 7{,}217 \\
\textsc{authentication} & refresh a token mid-session & 80.7 & 86.1 & 6{,}078 \\
\textsc{discovery} & pick among look-alikes & 76.1 & 80.5 & 6{,}932 \\
\bottomrule
\end{tabular}
\end{table}

\section{Per-Model Results on Every Task Set}
\label{app:full}

\begin{figure}[H]
\centering\includegraphics[width=0.5\textwidth]{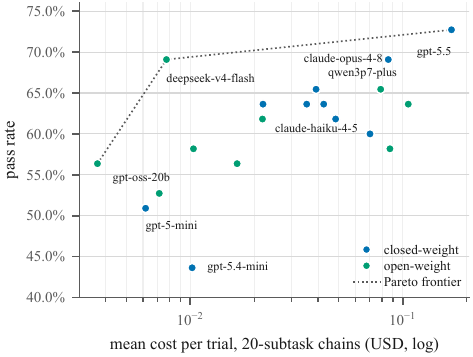}
\caption{Cost against 20-subtask pass rate. The frontier below \$0.01 per trial is entirely
open-weight. Prices come from the release-pinned table of July 19, 2026: a list-price comparison,
not a serving-efficiency claim.}
\label{fig:pareto}
\end{figure}

Table~\ref{tab:full} reports each model on all five task sets cut from the same worlds, together
with the 90\% cluster-bootstrap interval on the headline slice and the mean cost of each trial there.
Cost spans two orders of magnitude at comparable accuracy, which is a second reason not to read the
top of the 20-subtask column as a ranking.

\begin{table}[H]
\centering\small
\caption{All 19 models, frozen 1.0 bank, five epochs, temperature 1.0.
\textcolor{green1}{$\bullet$} marks open-weight models. ``All'' pools each task in the bank.
``Solo'' is the 241 single-subtask tasks. Chain-$k$ is the cumulative $k$-subtask slice. The
interval is a 90\% cluster bootstrap over worlds on the chain-20 slice. The \$/trial column is the mean cost on that same chain-20 slice.}
\label{tab:full}
\setlength{\tabcolsep}{4pt}
\begin{tabular}{@{}lrrrrrcr@{}}
\toprule
Model & All & Solo & Chain-5 & Chain-10 & Chain-15 & Chain-20 (90\% CI) & \$/trial \\
\midrule
gpt-5.5                        & 87.8 & 94.9 & 92.3 & 75.0 & 72.7 & \textbf{72.7} \;[45.5, 90.9] & 0.170 \\
claude-opus-4-8                & 88.6 & 96.0 & 92.3 & 68.3 & 72.7 & 69.1 \;[45.5, 90.9] & 0.086 \\
deepseek-v4-flash\,\textcolor{green1}{$\bullet$} & 88.4 & 95.6 & 98.5 & 75.0 & 70.9 & 69.1 \;[47.3, 89.1] & 0.008 \\
glm-5p2\,\textcolor{green1}{$\bullet$}           & 87.1 & 93.9 & 92.3 & 75.0 & 74.5 & 65.5 \;[40.0, 85.5] & 0.079 \\
qwen3p7-plus                   & 86.2 & 94.0 & 90.8 & 73.3 & 68.5 & 65.5 \;[40.0, 85.5] & 0.039 \\
claude-haiku-4-5               & 86.0 & 92.9 & 96.9 & 75.0 & 58.2 & 63.6 \;[36.4, 81.8] & 0.022 \\
deepseek-v4-pro\,\textcolor{green1}{$\bullet$}   & 86.7 & 94.6 & 92.3 & 75.0 & 65.5 & 63.6 \;[36.4, 81.8] & 0.106 \\
gpt-5.6-terra                  & 85.4 & 92.4 & 90.8 & 71.7 & 65.5 & 63.6 \;[36.4, 81.8] & 0.043 \\
qwen3p6-plus                   & 87.3 & 95.1 & 86.2 & 75.0 & 69.1 & 63.6 \;[40.0, 85.5] & 0.035 \\
claude-sonnet-4-6              & \textbf{89.0} & 95.7 & 86.2 & 75.0 & 69.1 & 61.8 \;[36.4, 83.6] & 0.048 \\
kimi-k2p7-code\,\textcolor{green1}{$\bullet$}    & 87.6 & 95.4 & 95.4 & 73.3 & 72.7 & 61.8 \;[36.4, 81.8] & 0.022 \\
claude-sonnet-5                & 87.8 & \textbf{96.5} & 84.6 & 73.3 & 72.7 & 60.0 \;[36.4, 81.8] & 0.070 \\
kimi-k2p6\,\textcolor{green1}{$\bullet$}         & 87.9 & 95.2 & 100.0 & 75.0 & 70.9 & 58.2 \;[34.5, 81.8] & 0.087 \\
minimax-m2p7\,\textcolor{green1}{$\bullet$}      & 85.5 & 93.0 & 92.3 & 75.0 & 61.8 & 58.2 \;[34.5, 80.0] & 0.010 \\
gpt-oss-20b\,\textcolor{green1}{$\bullet$}       & 80.9 & 88.2 & 86.2 & 73.3 & 61.8 & 56.4 \;[30.9, 81.8] & \textbf{0.004} \\
kimi-k2p5\,\textcolor{green1}{$\bullet$}         & 85.9 & 94.4 & 92.3 & 68.3 & 61.8 & 56.4 \;[30.9, 81.8] & 0.017 \\
gpt-oss-120b\,\textcolor{green1}{$\bullet$}      & 82.0 & 90.2 & 86.2 & 61.7 & 69.1 & 52.7 \;[27.3, 78.2] & 0.007 \\
gpt-5-mini                     & 80.6 & 89.6 & 83.1 & 58.3 & 56.4 & 50.9 \;[29.1, 72.7] & 0.006 \\
gpt-5.4-mini                   & 68.5 & 76.7 & 67.7 & 45.0 & 47.3 & 43.6 \;[23.6, 63.6] & 0.010 \\
\midrule
\emph{panel spread} & 20.5 & 19.8 & 32.3 & 30.0 & 27.3 & \emph{29.1} & --- \\ 
\bottomrule
\end{tabular}
\end{table}

\begin{figure}[H]
\centering\includegraphics[width=0.56\textwidth]{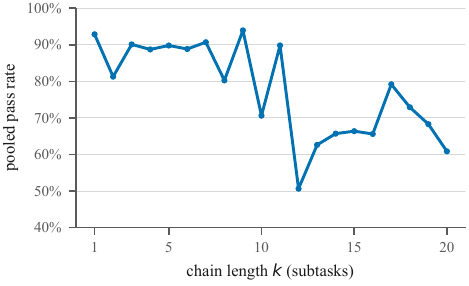}
\caption{Pass rate pooled over the 19-model panel against each chain length present in the
bank. The descent is real but not monotone: worlds contribute unevenly across lengths, per-length
samples are small, and the flagged cells of \S\ref{sec:defects} drag specific lengths
far down (per-cell zero; pooled minima near 50\%). The trend is more informative than individual per-length points.}
\label{fig:chainlen}
\end{figure}

\section{The 20-Subtask Slice by World}
\label{app:heat}

\begin{figure}[H]
\centering\includegraphics[width=0.88\textwidth]{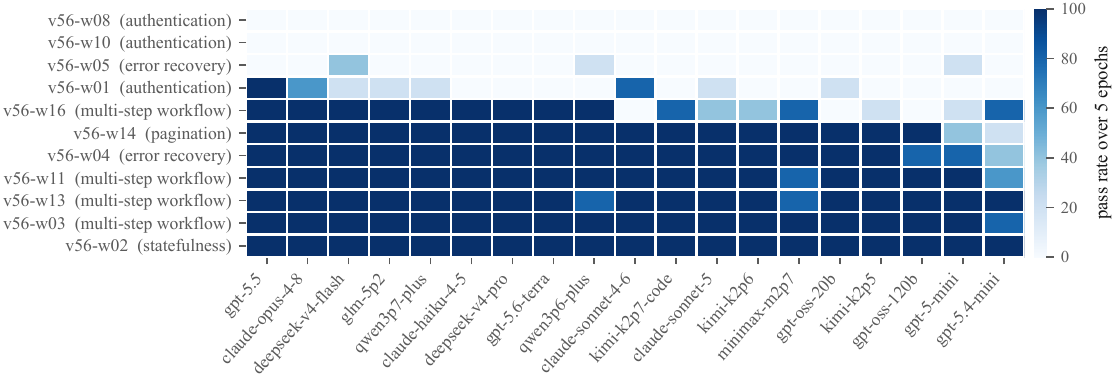}
\caption{The 20-subtask slice, with a cell for each (world, model) across five epochs, rows sorted
hardest-first and labeled with each world's dominant axis. Variation across worlds (rows) exceeds variation across models (columns). The two rows
at zero are cells no model passes for which a reference solution nonetheless clears the assembled grader (\S\ref{sec:defects}); whether they are hard or a compression artifact is left open. This implies that
an all-zero row must be diagnosed before it is interpreted.}
\label{fig:heatmap}
\end{figure}

\section{The Adversarial Grader Audit in Full}
\label{app:audit}

\begin{table}[H]
\centering\small
\caption{The six exploits an independent red-team pass found in the finished grader, and their
fixes. All numbers reported in this work are obtained after these fixes.}
\label{tab:audit}
\begin{tabular}{@{}llp{5.3cm}p{4.4cm}@{}}
\toprule
Axis & Class & The exploit & The fix \\
\midrule
\textsc{schema} & false neg. & a creation \texttt{POST} returns 201 while the check pinned 200, rejecting the minimal correct solve & accept the real success code (\texttt{[200,201]}) \\
\textsc{schema} & false pos. & the graded value was echoable (present in a payload the agent could copy without fetching it) & remove the echo; pin a distinct canary the mock serves only on the real path \\
\textsc{statefulness} & false neg. & a currency comparator demanded a \texttt{USD} unit the agent had no source for, rejecting correct numbers & grade numerically; stop forcing a hallucinated unit \\
\textsc{pagination} & false pos. & the total was also served on a summary endpoint, making the task passable without paging & remove the leak; a terminal-page marker forces the full walk \\
\textsc{multistep} & false pos. & the final canary was reachable by guessing a single URL, skipping the chain & gate the value behind stateful mock logic and a non-guessable id \\
\textsc{multistep} & false pos. & a value was readable \emph{before} the publish step that was supposed to produce it & gate the read on the publish; add chain-level state checks \\
\bottomrule
\end{tabular}
\end{table}

\section{Datasheet, Licensing, and Reproduction}
\label{app:datasheet}

\paragraph{Composition and collection.} The bank is 467 tasks in 13 procedurally generated
worlds (i.e., 241 solo subtasks and 226 cumulative chains). The bank involves no human subjects, no user
traffic, and no personally identifiable information. Each world, mock backend, instruction, and
expected answer is synthesized by the pipeline of \S\ref{sec:building}. Task instructions are English.

\paragraph{Artifact URLs.} Harness and frozen bank:
\anonurl{https://github.com/postmanlabs/APIFlow-Bench}{\nolinkurl{github.com/postmanlabs/APIFlow-Bench}};
transcripts:
\anonurl{https://github.com/postmanlabs/apiflow-bench-transcripts}{\nolinkurl{github.com/postmanlabs/apiflow-bench-transcripts}};
leaderboard: \anonurl{https://www.postman.com/ai/apiflow-leaderboard/overview/}{\nolinkurl{postman.com/ai/apiflow-leaderboard}}.

\paragraph{Licensing and distribution.} The harness, bank, scoring code, and leaderboard builder are
released under Apache-2.0 at \anonurl{https://github.com/postmanlabs/APIFlow-Bench}{\nolinkurl{github.com/postmanlabs/APIFlow-Bench}}. All 44{,}362 raw
trial transcripts are released at
\anonurl{https://github.com/postmanlabs/apiflow-bench-transcripts}{\nolinkurl{github.com/postmanlabs/apiflow-bench-transcripts}}, and model-level bundles are attached
to the repository's releases. The authors bear responsibility for the artifacts and confirm the
license permits the stated use.

\paragraph{Reproduction without API keys.} \texttt{scripts/bank\_sha256.py} verifies the bank
byte for byte against the content hash pinned in \texttt{registry.json}.
\texttt{scripts/golden\_replay.py} replays each world's reference trajectory forward and in its
sabotaged form. Both run against the local mocks with no network access and no credentials.
Reproducing the leaderboard requires provider API keys. On the published panel the full
$467 \times 5$ grid cost between roughly \$8 and \$730 \emph{in total} for each model under the
release-pinned price table. These totals differ in scale from the cost column of
Table~\ref{tab:full}. This is because the cost column reports the mean cost of each trial on the
chain-20 slice.

\paragraph{Maintenance and versioning.} Each release is pinned in \texttt{registry.json} by bank
path, content hash, epoch count, and price-table version. Hence, running 1.0 in a later year reproduces
1.0 as the repository grows. The pinned hash covers the annotation-stripped bank that ships publicly.
Golden replay re-verifies equivalence to the evaluated bank after the strip. The flagged cells are listed in the repository, not silently
repaired in a frozen bank (\S\ref{sec:defects}). A refreshed bank with a private held-out split is
planned to anchor rankings once the public bank has aged.

\paragraph{Safe use.} Task code (\texttt{validator.py}, \texttt{mock\_overrides.py}) executes in the
harness process. Hence, third-party task banks should be treated like other executed code.

\end{document}

%% file: fig-overview.tex
\begin{tikzpicture}[
  font=\small,
  box/.style={draw=blkgray!50, rounded corners=2pt, line width=.5pt, fill=white,
              align=left, inner sep=3.5pt},
  hdr/.style={font=\small\bfseries, inner sep=0pt, anchor=south west},
  tool/.style={draw=blue1!70, fill=blue1!8, rounded corners=1.2pt, line width=.4pt,
               inner sep=1.8pt, font=\scriptsize\ttfamily},
  flow/.style={-{Stealth[length=3.4pt,width=2.8pt]}, draw=blkgray!75, line width=.6pt},
  note/.style={font=\scriptsize, text=blkgray!85, align=left, inner sep=0pt},
]
\def\colw{40mm}
\def\taskw{52mm}

\node[box, text width=\taskw, anchor=north west] (task) at (0,0) {%
  {\ttfamily\scriptsize\textcolor{blkgray!85}{My nightly pull 401s on the readings}}\\[-2pt]
  {\ttfamily\scriptsize\textcolor{blkgray!85}{API. This is the broken call:}}\\[3pt]
  {\ttfamily\scriptsize GET /v3/stations/stn-12/readings}\\[-2pt]
  {\ttfamily\scriptsize Authorization: Bearer tk\_live\_OLD}\\[3pt]
  {\ttfamily\scriptsize\textcolor{orange1}{\{"error":"token\_expired", "hint":}}\\[-2pt]
  {\ttfamily\scriptsize\textcolor{orange1}{ "refresh via POST /v3/oauth/token"\}}}\\[3pt]
  {\ttfamily\scriptsize\textcolor{blkgray!85}{Refresh auth and get the live reading}}\\[-2pt]
  {\ttfamily\scriptsize\textcolor{blkgray!85}{for stn-12: the value and the ref.}}%
};
\node[hdr] at ([yshift=1.5pt]task.north west) {(a)~A failure-first task};
\node[note, anchor=north west, text width=\taskw] at ([yshift=-2.5pt]task.south west)
  {axis \textsc{authentication} $\cdot$ difficulty \textsc{hard}\\ canary GUID in the
   \texttt{task.yaml} header};

\node[box, text width=\colw, anchor=north west, minimum height=29mm]
  (arena) at ([xshift=7mm]task.north east) {\strut};
\node[hdr] at ([yshift=1.5pt]arena.north west) {(b)~A neutral seven-tool scaffold};

\node[note, anchor=north west] at ([shift={(3.5pt,-3.5pt)}]arena.north west)
  {agent $=$ Inspect AI \texttt{react()} loop;};
\node[note, anchor=north west] at ([shift={(3.5pt,-11pt)}]arena.north west)
  {no planner, summariser or retry logic.};

\node[tool, anchor=north west] (t1) at ([shift={(3.5pt,-21pt)}]arena.north west) {read};
\node[tool, right=1.6pt of t1] (t2) {write};
\node[tool, right=1.6pt of t2] (t3) {edit};
\node[tool, right=1.6pt of t3] (t4) {search};
\node[tool, below=1.8pt of t1.south west, anchor=north west] (t5) {execute};
\node[tool, right=1.6pt of t5] (t6) {clarify};
\node[tool, below=1.8pt of t5.south west, anchor=north west] (t7) {report\_blocked};
\node[note, anchor=north west] at ([shift={(0,-5.5pt)}]t7.south west)
  {polymorphic over five entity kinds:};
\node[note, anchor=north west] at ([shift={(0,-14pt)}]t7.south west)
  {\hspace*{0pt}{\ttfamily request response variable}\\ {\ttfamily spec handle}};

\node[box, fill=blue1!6, draw=blue1!45, text width=\colw, anchor=north, align=center]
  (mock) at ([yshift=-6mm]arena.south) {%
  \textbf{deterministic mock REST backend}\\[2pt]
  {\scriptsize expiring tokens $\cdot$ partial writes\\[-1pt]
   429 with \texttt{Retry-After} $\cdot$ paging\\[-1pt]
   deprecated \texttt{/v1} decoys}\\[3pt]
  {\scriptsize it mints the \textcolor{orange1}{\textbf{canary value}} only on the correct path}};
\draw[flow, <->] (arena.south) -- (mock.north);

\node[box, text width=\colw, anchor=north west] (g1) at ([xshift=7mm]arena.north east) {\hyphenpenalty=10000\relax%
  \textbf{G1 $\cdot$ state}---\emph{the work happened}\\[2.5pt]
  {\scriptsize The earned canary is in the final workspace, and a data-flow check ties it back to
   the response it had to come from.}};
\node[hdr] at ([yshift=1.5pt]g1.north west) {(c)~Grading is G1 $\wedge$ G2};

\node[box, text width=\colw, anchor=north west] (g2) at ([yshift=-7mm]g1.south west) {\hyphenpenalty=10000\relax%
  \textbf{G2 $\cdot$ answer}---\emph{the user was served}\\[2.5pt]
  {\scriptsize A typed answer card, compared field by field (numeric / currency / string / set),
   never by substring.}};
\node[font=\bfseries, text=blkgray] at ($(g1.south)!.5!(g2.north)$) {$\wedge$};

\node[box, fill=green1!12, draw=green1!60, text width=\colw, anchor=north west, align=center]
  (verdict) at ([yshift=-4.5mm]g2.south west) {\textbf{PASS}---both surfaces,\\ or nothing};
\draw[flow] (g2.south) -- (verdict.north);

\draw[flow] (task.east) -- node[above, font=\scriptsize, pos=.45] {prompt} (arena.west);
\draw[flow] (arena.east) -- node[above, font=\scriptsize] {final} node[below, font=\scriptsize] {state} (g1.west);
\end{tikzpicture}

%% file: fig-pipeline.tex
\begin{tikzpicture}[
  font=\small, rounded corners=2pt,
  cell/.style={draw=blkgray!55, rounded corners=2pt, line width=.5pt, fill=white,
               minimum width=13.5mm, minimum height=7.5mm, align=center, inner sep=1.5pt},
  st/.style={draw=blue1!60, fill=blue1!10, rounded corners=1.5pt, line width=.45pt,
             minimum width=6.5mm, minimum height=5.5mm, inner sep=1pt, font=\scriptsize},
  pstep/.style={draw=blkgray!55, rounded corners=2pt, line width=.5pt, fill=white,
                align=center, inner sep=3pt, font=\scriptsize, text width=21mm},
  gate/.style={pstep, draw=orange1!75, fill=orange1!8},
  pout/.style={pstep, draw=green1!65, fill=green1!10},
  flow/.style={-{Stealth[length=3.4pt,width=2.8pt]}, draw=blkgray!75, line width=.6pt},
  back/.style={-{Stealth[length=3.2pt,width=2.6pt]}, draw=orange1, line width=.55pt,
               dash pattern=on 1.7pt off 1.4pt},
  hdr/.style={font=\small\bfseries, inner sep=0pt, anchor=south west},
  note/.style={font=\scriptsize, text=blkgray!85, align=left, inner sep=0pt},
]

\node[st, anchor=north west] (s0) at (0,0) {$S_0$};
\node[cell, right=3mm of s0] (c1) {subtask 1\\[-1.5pt]{\scriptsize\itshape pagination}};
\node[st,   right=3mm of c1] (s1) {$S_1$};
\node[cell, right=3mm of s1] (c2) {subtask 2\\[-1.5pt]{\scriptsize\itshape schema}};
\node[st,   right=3mm of c2] (s2) {$S_2$};
\node[cell, right=3mm of s2] (c3) {subtask 3\\[-1.5pt]{\scriptsize\itshape auth}};
\node[right=3mm of c3, font=\small, inner sep=1pt] (dots) {$\cdots$};
\node[cell, right=3mm of dots] (ck) {subtask 20\\[-1.5pt]{\scriptsize\itshape statefulness}};
\node[st,   right=3mm of ck] (sk) {$S_{20}$};
\foreach \a/\b in {s0/c1,c1/s1,s1/c2,c2/s2,s2/c3,c3/dots,dots/ck,ck/sk}{\draw[flow] (\a)--(\b);}

\draw[back] (c1.north) -- ++(0,5mm) -| (c3.north);
\node[note, text=orange1, anchor=south] at ([yshift=5.4mm]$(c1.north)!.5!(c3.north)$)
  {\itshape coupling ledger: subtask 1 plants an id, subtask 3 must consume it};
\node[hdr] at ([yshift=11mm]s0.north west) {(a)~One world, grown forward};
\node[note, anchor=north west, text width=118mm] at ([yshift=-2.5mm]s0.south west)
  {$S_i$ = the agent workspace plus the mock's accumulated side effects, carried forward from a run
   that \emph{passed}.};

\node[pstep, anchor=north west] (p1) at ([yshift=-24mm]s0.south west)
  {propose a subtask\\ seeded from $S_{i-1}$};
\node[pstep, right=7mm of p1] (p2) {write the mock overrides\\ and the validator};
\node[gate,  right=7mm of p2] (p3) {\textbf{self-test triad}\\[1pt]
  blank~\ding{55}\quad key~\ding{51}\\ sabotage~\ding{55}};
\node[gate,  right=7mm of p3] (p4) {\textbf{solvability gate}\\[1pt]
  the oracle must reach pass@10 $\ge 3$};
\node[pout,  right=7mm of p4] (p5) {\textbf{accept}\\ save $S_i$, settle any coupling};
\foreach \a/\b in {p1/p2,p2/p3,p3/p4,p4/p5}{\draw[flow] (\a)--(\b);}

\draw[back] (p3.north) -- ++(0,3.5mm) -| (p2.north);
\node[note, text=orange1, anchor=south] at ([yshift=3.9mm]$(p2.north)!.5!(p3.north)$)
  {redesign the validator once};
\draw[back] (p4.south) -- ++(0,-4.5mm) -| (p1.south);
\node[note, text=orange1, anchor=north] at ([yshift=-6.2mm]$(p1.south)!.5!(p4.south)$)
  {repair in place ($\le 3$), re-propose ($\le 1$), else truncate the chain here};
\node[hdr] at ([yshift=10mm]p1.north west)
  {(b)~Inside one cell: two gates clear before any model is scored};

\node[pstep, anchor=north west, text width=27mm] (f1) at ([yshift=-25mm]p1.south west)
  {cut the accepted chain at every $k$: subtasks $1, \ldots, k$ freeze into one \texttt{chain-1to$k$} task};
\node[pstep, anchor=north west, text width=27mm] (f2) at ([yshift=-4mm]f1.south west)
  {each segment also freezes on its own into a \texttt{solo} task};
\coordinate (mid) at ($(f1.east)!.5!(f2.east)$);
\node[gate, anchor=west, text width=31mm] (f3) at ([xshift=13mm]mid)
  {\textbf{golden replay}\\[1pt] forward replay must \textsc{pass};\\
   sabotaged replay must \textsc{fail}};
\node[pout, right=8mm of f3, text width=29mm] (f4)
  {\textbf{frozen release}\\[1pt] 467 tasks over 13 worlds,\\ sha256-pinned in \texttt{registry.json}};
\draw[flow] (f1.east) -| ($(f1.east)!.5!(f3.west |- f1.east)$) |- (f3.west);
\draw[flow] (f2.east) -| ($(f2.east)!.5!(f3.west |- f2.east)$) |- (f3.west);
\draw[flow] (f3.east) -- (f4.west);
\node[hdr] at ([yshift=2mm]f1.north west) {(c)~Freezing an accepted chain into the shipped bank};
\end{tikzpicture}